\pdfoutput=1

\documentclass[11pt]{article}

\usepackage{acl}
\usepackage{graphicx}
\usepackage{times}
\usepackage{latexsym}
\usepackage{makecell}

\usepackage[T1]{fontenc}

\usepackage[utf8]{inputenc}

\usepackage{microtype}

\usepackage{booktabs}

\newcommand{\BLEURT}{\textsc{Bleurt}}
\newcommand{\COMET}{\textsc{Comet}}
\newcommand{\BLEURText}{\textsc{Bleurt-Extended}}

\title{Learning Compact Metrics for MT}

\author{Amy Pu\thanks{ \, Work done during an internship at Google.}  \ \ \ Hyung Won Chung \thanks{ \, Work done as a Google AI resident.} \ \ \ Ankur P. Parikh \ \ \ Sebastian Gehrmann  \ \ \  Thibault Sellam\\
Google Research \\
New York, NY \\
\texttt{amy\_pu@alumni.brown.edu} \\
\texttt{\{hwchung, aparikh, gehrmann, tsellam \}@google.com}
}

\begin{document}
\maketitle

\begin{abstract}

Recent developments in machine translation and multilingual text generation have led researchers to adopt trained metrics such as COMET or BLEURT, which treat evaluation as a regression problem and use representations from multilingual pre-trained models such as XLM-RoBERTa or mBERT. Yet studies on related tasks suggest that these models are most efficient when they are large, which is costly and impractical for evaluation. We investigate the trade-off between multilinguality and model capacity with RemBERT, a state-of-the-art multilingual language model, using data from the WMT Metrics Shared Task. We present a series of experiments which show that model size is indeed a bottleneck for cross-lingual transfer, then demonstrate how distillation can help addressing this bottleneck, by leveraging synthetic data generation and transferring knowledge from one teacher to multiple students trained on related languages. Our method yields up to 10.5\% improvement over vanilla fine-tuning and reaches 92.6\% of RemBERT’s performance using only a third of its parameters.\looseness=-1

\end{abstract}

\section{Introduction}

Recent improvements in Machine Translation (MT) and multilingual Natural Language Generation (NLG) have led researchers to question the use of n-gram overlap metrics such as BLEU and ROUGE~\citep{papineni2002bleu,lin2004rouge}. Since these metrics focus solely on surface-level aspects of the generated text, they correlate poorly with human evaluation, especially when models are producing high-quality text~\cite{belz2006comparing,callison2006re,ma2019results,mathur2020tangled}. This has led to a surge of interest in \emph{learned} metrics that cast evaluation as a regression problem and leverage \emph{pre-trained multilingual models} to capture the semantic similarity between references and generated text~\cite{celikyilmaz2020evaluation}. Popular examples of those metrics include \COMET~\cite{rei2020comet} and \BLEURText{}~\cite{sellam2020bleurt}, based on XLM-RoBERTa~\cite{conneau2019cross, conneau2019unsupervised} and mBERT~\cite{devlin2018bert} respectively. These metrics deliver superior performance over those based on lexical overlap, outperforming even crowd-sourced annotations~\cite{freitag2021experts, mathur2020results}.

Large pre-trained models benefit learned metrics in at least two ways. First, they allow for \emph{cross-task transfer}: the contextual embeddings they produce allow researchers to address the relative scarcity of training data that exist for the task, especially with large models such as BERT or XLNet~\cite{zhang2019bertscore, devlin2018bert, yang2019xlnet}. Second, they allow for \emph{cross-lingual transfer}:  MT evaluation is often multilingual, yet few, if any, popular datasets cover more than 20 languages.  Evidence suggests that training on many languages improves performance on languages for which there is little training data, including the \emph{zero-shot} setup, in which no fine-tuning data is available~\cite{conneau2019cross, sellam2020learning, conneau2018xnli, pires2019multilingual}. 

However, the accuracy gains only appear if the model is large enough. In the case of cross-lingual transfer, this phenomenon is known as \emph{the curse of multilinguality}: to allow for positive transfer, the model must be scaled up with the number of languages~\cite{conneau2019cross}. Scaling up metric models is particularly problematic, since they must often run alongside an already large MT or NLG model and, therefore, must share hardware resources (see ~\citet{shu2021reward} for a recent use case). This contention may lead to impractical delays, it increases the cost of running experiments, and it prevents researchers with limited resources from engaging in shared tasks.

We first present a series of experiments that validate that previous findings on cross-lingual transfer and the curse of multilinguality apply to the metrics domain, using RemBERT (Rebalanced mBERT), a multilingual extension of BERT~\cite{chung2020rethinking}. We then investigate how a combination of multilingual data generation and distillation can help us reap the benefits of multiple languages while keeping the models compact. Distillation has been shown to successfully transfer knowledge from large models to smaller ones~\cite{hinton2015distilling}, but it requires access to a large corpus of unlabelled data~\cite{sanh2019distilbert,turc2019well}, which does not exist for our task. Inspired by~\citet{sellam2020bleurt}, we introduce a data generation method based on random perturbations that allows us to synthesize arbitrary amounts of multilingual training data. We generate an 80M-sentence distillation corpus in 13 languages from Wikipedia, and show that we can improve a vanilla pre-trained distillation setup~\cite{turc2019well} by up to 12\%. A second, less explored benefit of distillation is that it lets us partially bypass the curse of multilinguality. Once the teacher (i.e., larger) model has been trained, we can generate training data for any language, including the zero-shot ones. Thus, we are less reliant on cross-lingual transfer. We can lift the restriction that one model must carry all the languages, and train smaller models, targeted towards specific language families. Doing so increases performance further by up to~4\%. Combining these two methods, we match 92.6\% of the the largest RemBERT model's performance using only a third of its parameters.

A selection of code and models is available online at \url{https://github.com/google-research/bleurt}.

\section{Multilinguality and Model Size}
\label{sec:motivation}

To motivate our work, we quantify the trade-off between multilinguality and model capacity using data from the WMT Shared Task 2020, the most recent benchmark for MT evaluation metrics. The phenomenon has been well-studied for tasks such as translation~\citep{aharoni2019massively} and language inference~\citep{conneau-etal-2020-unsupervised},
but it is less well understood in the context of evaluation metrics.

\paragraph{Task and Data} In the WMT Metrics task, participants evaluate the quality of MT systems with automatic metrics for 18 language pairs---10 to-English, 8 from-English. The success criterion is correlation with human ratings.\footnote{The study focuses on segment-level correlation but we also report systems-level results in the appendix.} Following established approaches~\cite{ma2018results, ma2019results}, we utilize the human ratings from the previous years' shared tasks for training. Our training set contains 479k triplets \texttt{(Reference translation, MT output,  Rating)} in 12 languages, and it is heavily skewed towards English. It covers the target languages of the benchmark except Polish, Tamil, Japanese and Inuktitut.\footnote{The target languages are: English, Czech, German, Japanese, Polish, Russian, Tamil, Chinese, Inuktitut. We train on English, German, Chinese, Czech, Russian, Finnish, Estonian, Kazakh, Lithuanian, Gujarati, French, and Turkish.} We evaluate the first three in a \emph{zero-shot} fashion and do no report results on Inuktitut because its alphabet is not covered by RemBERT.

\begin{figure}[t!]
    \centering
    \includegraphics[width=\columnwidth]{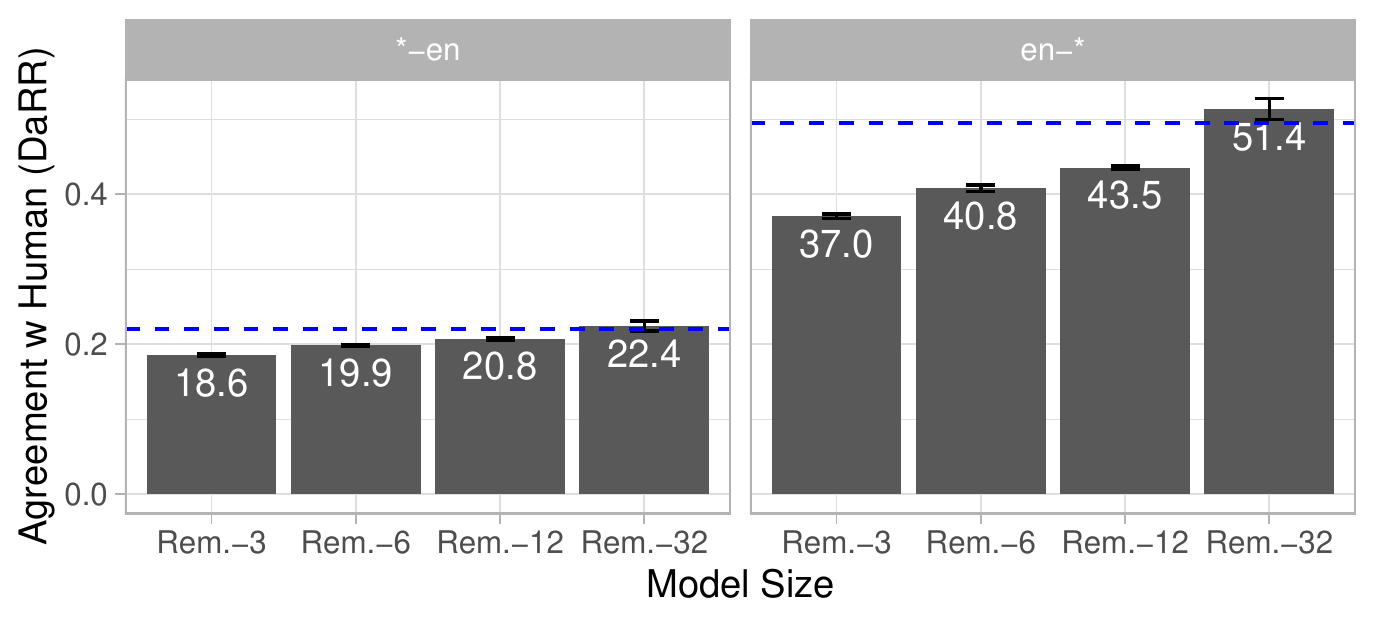}
    \caption{Performance of our models. The \textcolor{darkblue}{dashed line} represents the performance of \BLEURT{}-extended~\cite{sellam2020learning}. The metric is WMT Metrics DaRR~\cite{mathur2020results}, a robust variant of Kendall Tau, higher is better. We run each experiment with 5 random seeds, report the mean result and Normal-based 95\% confidence intervals.}
    \label{fig:model_sizes}
\end{figure}

\paragraph{Models} Like \COMET~\cite{rei2020comet} and \BLEURT~\cite{sellam2020bleurt}, we treat evaluation as a regression problem where, given a reference translation $\mathbf{x}$ (typically produced by a human) and predicted translation $\tilde{\mathbf{x}}$ (produced by an MT system), the goal is to predict a real-valued human rating $y$. As is typical, we leverage pretrained representations~\citep{peters2018deep} to achieve strong performance. Specifically, we first embed sentence pairs into a fixed-width vector $\mathbf{v} = \mathbf{F}(\mathbf{x}, \tilde{\mathbf{x}})$ using a pretrained model $\mathbf{F}$ and use this vector as input to a linear layer:
$\hat{y} = \mathbf{W} \mathbf{v} + \mathbf{b}$, where $\mathbf{W}$ and $\mathbf{b}$ are the weight matrix and bias vector respectively.

For the pretrained model $\mathbf{F}$, we use RemBERT~\cite{chung2020rethinking}, a recently published extension of mBERT~\cite{devlin2018bert} pre-trained on 104 languages using a combination of Wikipedia and  mC4~\citep{DBLP:journals/jmlr/RaffelSRLNMZLL20}. Because RemBERT is massive (32 layers, 579M parameters during fine-tuning) we pre-trained three smaller variants, RemBERT-3, RemBERT-6, and RemBERT-12, using Wikipedia data in 104 languages. The models are respectively 95\%, 92\%, and 71\% smaller, with only 3, 6, and 12 layers. We refer to RemBERT as RemBERT-32 for consistency.  The details of architecture, pre-training and fine-tuning are in the appendix.

Figure~\ref{fig:model_sizes} presents the performance of the models. RemBERT-32 is on par with \BLEURText, a metric based on a similar model which performed well at WMT Metrics 2020.\footnote{Model provided by the authors. The results diverge from~\citet{mathur2020results} on \texttt{en-zh}, for which they submitted a separate metric but the conclusions are similar.} It also corroborates that for a fixed number of languages, larger models  perform better.
 
\begin{figure}[t!]
    \centering
    \includegraphics[width=\columnwidth]{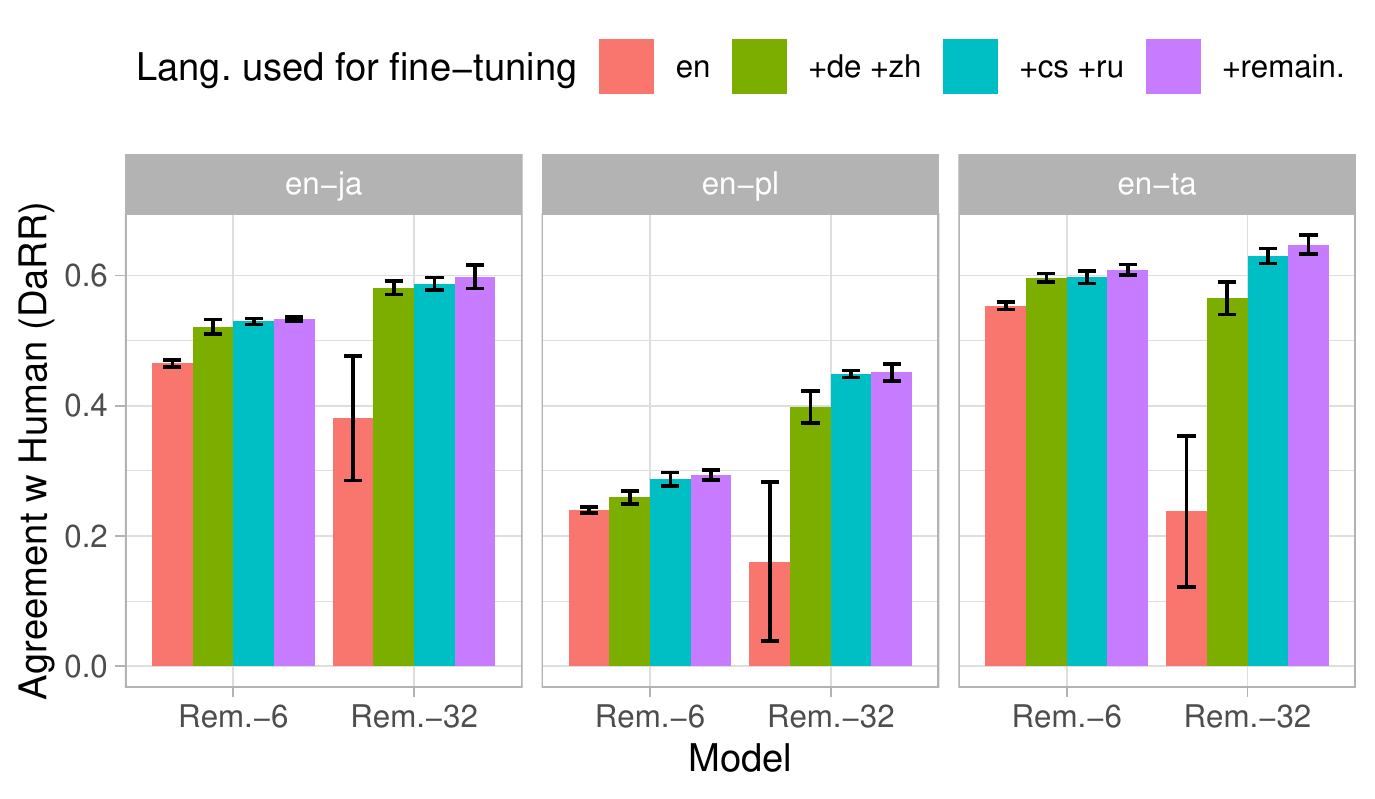}
    \caption{Impact of the number of fine-tuning languages on zero-shot performance, using RemBERT-6 and RemBERT-32 on \texttt{en-ja}, \texttt{en-pl}, and \texttt{en-ta}.}
    \label{fig:finetuning}
\end{figure}

\paragraph{Cross-lingual transfer during fine-tuning} Figure~\ref{fig:finetuning} displays the performance of RemBERT-6 and RemBERT-32 on the zero-shot languages as we increase the number of languages used for fine-tuning. We start with English, then add the languages cumulatively, in decreasing order of frequency (without adding data for any of the target languages). Cross-lingual transfer works: in all cases, adding languages improves performance. The effect is milder on RemBERT-6, which consistently starts higher but finishes lower. The appendix presents additional details and results.




\begin{figure}[t!]
    \centering
    \includegraphics[width=\columnwidth]{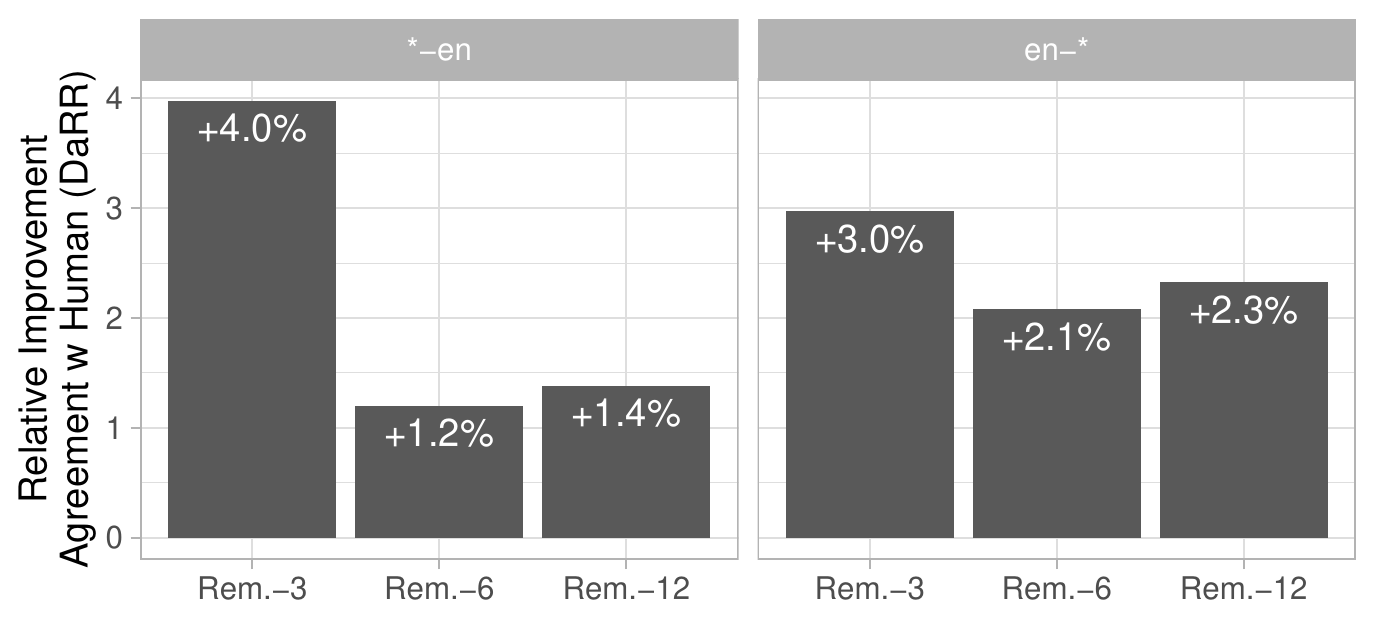}
    \caption{Improvement after removing 86 languages from from pre-training. y-axis: relative performance improvement over a RemBERT of equal size pretrained on 104 languages. Additional details in the appendix.}
    \label{fig:pretraining}
\end{figure}

\paragraph{Capacity bottleneck in pre-training} To further understand the effect of multilinguality, we pre-trained the smaller models from scratch using 18 languages of WMT instead of 104, and fine-tuned on the whole dataset. Figure~\ref{fig:pretraining} presents the results: performance increases in all cases, especially for RemBERT-3. This suggests that the models are at capacity and that the 100+ languages of the pre-training corpus compete with one another.

\paragraph{Takeaways} Learned metrics are subject to conflicting requirements. On one hand, the opportunities offered by pre-training and cross-lingual transfer encourage researchers to use large, multilingual models. On the other hand, the limited hardware resources inherent to evaluation call for smaller models, which cannot easily keep up with massively multilingual pre-training. We address this conflict with distillation.

\section{Addressing the Capacity Bottleneck}
\label{sec:distillation}

\begin{figure}[t!]
    \centering
    \includegraphics[width=.9\columnwidth]{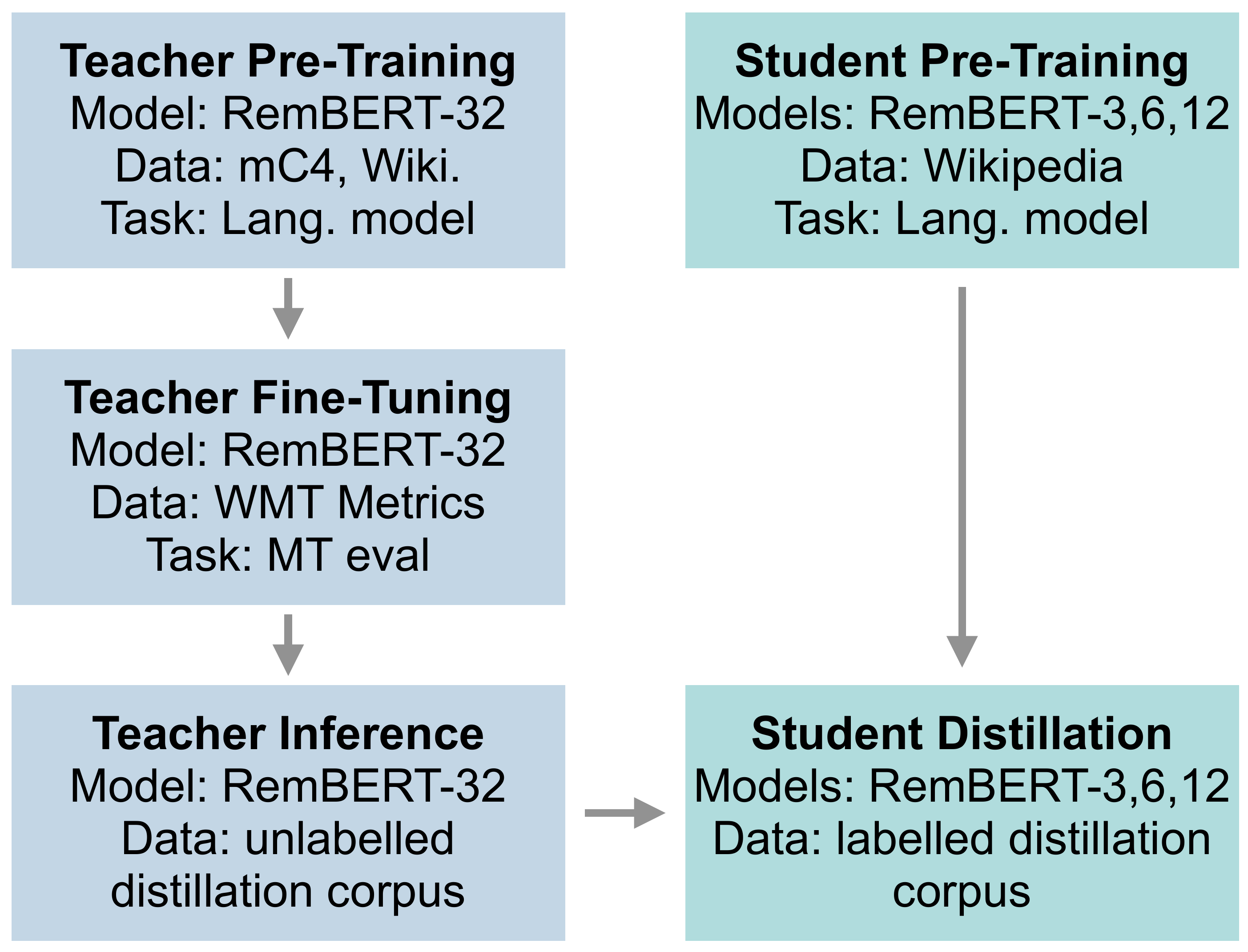}
    \caption{Overview of the default distillation pipeline.}
    \label{fig:distillation}
\end{figure}

\begin{table*}[!t]
\centering
\small
\begin{tabular} { l l | c  c |  c c c c c c c c  }
\toprule
Model & & *-en & en-* & en-cs & en-de & en-ja & en-pl & en-ru & en-ta & en-zh\\
\midrule
\multicolumn{2}{l |}{\textsc{Comet}~\cite{rei2020comet}$^\dagger{}$ - \textit{550M params}}& - & 52.4 & 66.8 & 46.8 & 62.4 & 46.2 & 34.4 & 67.1 & 43.2 \\
\multicolumn{2}{l |}{\textsc{Prism}~\cite{thompson-post-2020-automatic}$^\dagger{}$ - \textit{745M pa.}} & - & 45.5 & 61.9 & 44.7 & 57.9 &41.4 & 28.3 &44.8 & 39.7 \\
\multicolumn{2}{l |}{\BLEURT{}-Ext.~\cite{sellam2020learning} - \textit{425M pa.}} & 22.0 & 49.8 & 68.8 & 44.7 & 53.3 & 43.0 & 30.6 & 64.3 & 44.2\\
\midrule
\multicolumn{2}{l |}{Teacher: RemBERT-32 - \textit{579M params}} & 22.5 & 52.3 & 69.3 & 45.9 & 61.7 & 45.4 & 31.0 & 66.6 & 45.9\\
\midrule
RemBERT-3 & Fine-tuning & 18.5 & 36.9 & 42.8 & 33.0 & 49.7 & 26.2 & 16.0 & 57.4 & 33.1\\
\textit{30M params} & Distill WMT & 16.3 & 34.8 & 43.3 & 29.0 & 46.8 & 22.0 & 15.4 & 56.1 & 31.3\\
 & Distill WMT+Wiki & 19.1 & 39.1 & 42.3 & \textbf{34.4} & 53.6 & 26.9 & 18.9 & \textbf{60.3} & 37.6\\
 & 1-to-N distill &\textbf{19.9} & \textbf{40.1} & \textbf{47.3} & 32.9 & \textbf{54.4} & \textbf{27.3} & \textbf{19.3} & 60.0 & \textbf{39.6}\\
\midrule
RemBERT-6 & Fine-tuning & 19.6 & 40.3 & 51.4 & 35.0 & 53.6 & 28.5 & 19.0 & 60.2 & 34.8\\
\textit{45M params} & Distill WMT & 19.9 & 40.4 & 53.1 & 34.8 & 52.1 & 28.4 & 17.9 & 60.1 & 36.3\\
 & Distill WMT+Wiki & 20.7 & 42.6 & 51.6 & 36.7 & 55.6 & 30.2 & 20.3 & \textbf{63.1} & 40.9\\
 & 1-to-N Distill & \textbf{21.0} & \textbf{44.4} & \textbf{56.1} & \textbf{38.3} & \textbf{57.1} & \textbf{34.6} & \textbf{22.2} & 59.9 & \textbf{42.9}\\
\midrule
RemBERT-12 & Fine-tuning & 20.6 & 43.8 & 57.4 & 36.7 & 56.1 & 33.0 & 23.4 & 62.2 & 37.5\\
\textit{167M params} & Distill WMT & 21.4 & 44.8 & 59.3 & 39.3 & 56.0 & 34.7 & 22.9 & 63.6 & 38.1\\
 & Distill WMT+Wiki & \textbf{21.9} & 47.3 & 59.2 & \textbf{40.8} & \textbf{57.9} & 37.4 & 26.4 & \textbf{65.3} & \textbf{44.2}\\
 & 1-to-N Distill & 21.7 & \textbf{48.4} & \textbf{64.2} & 40.2 & 57.6 & \textbf{41.3} & \textbf{28.4} & 63.7 & 43.5\\

\bottomrule
\end{tabular}
\caption{Segment-level agreement with human ratings of our distillation setups. The metric is WMT Metrics DaRR \cite{mathur2020results}, a robust variant of Kendall Tau, higher is better. The dagger$^\dagger{}$ indicates that the results were obtained from the WMT report. We omit \texttt{*-en} for these because of inconsistencies between the benchmark implementations. The number of parameters describes the size of the pre-trained model (without the terminal fully connected layer). For \textsc{Comet}, we reported the parameter count of XLM-RoBERTa-large~\cite{conneau2019unsupervised}, mentioned in~\citet{rei-etal-2020-unbabels}. The appendix presents additional details, baselines and systems-level results.}
\label{table:distillation-results}
\end{table*}

The main idea behind distillation is to train a small model (the \emph{student}) on the output of larger one (the \emph{teacher})~\cite{hinton2015distilling}.
This technique is believed to yield better results than training the smaller model directly on the end task because the teacher can provide pseudo-labels for an arbitrary large collection of training examples. Additionally, \citet{turc2019well} have shown that pre-training the student on a language model task before distillation improves its accuracy (in the monolingual setting), a technique known as \emph{pre-trained distillation}.

Since pre-trained distillation was shown to be simple and efficient, we use it for our base setup. Figure~\ref{fig:distillation} summarizes the steps: we fine-tune RemBERT-32 on human ratings, run it on an unlabelled distillation corpus, and use the predictions to supervise RemBERT-3, 6, or 12. By default, we use the WMT corpus for distillation, i.e., we use the same sentence pairs for teacher fine-tuning and student distillation (but with different labels).

\paragraph{Improvement 1: data generation} Distillation requires access to a large multilingual dataset of sentence pairs \texttt{(reference, MT output)} to be annotated by the teacher. Yet the WMT Metrics corpus is relatively small, and no larger corpus exists in the public domain. To address this challenge we generate pseudo-translations by perturbing sentences from Wikipedia. We experiment with three types of perturbations: back-translation, word substitutions with mBERT, and random deletions. The motivation is to generate surface-level noise and paraphrases, to expose the student to the different types of perturbations that an MT system could introduce. In total, we generate 80 million sentence pairs in 13 languages. The approach is similar to~\citet{sellam2020bleurt}, who use perturbations to generate pre-training data in English. We present the details of the approach in the appendix.

\paragraph{Improvement 2: 1-to-N distillation}
Another benefit of distillation is that it allows us to lift the constraint that one model must carry all the languages. In a regular fine-tuning setup, it is necessary to pack as many languages as possible in the same model because training data is sparse or non-existent in most languages. In our distillation setup, we can generate vast amounts of data for any language of Wikipedia. It is thus possible to bypass the capacity constraint by training $N$ specialized students, focused on a smaller number of languages. For our experiments, we pre-train five versions of each RemBERT, which cover between 3 and 18 languages each. We tried to form clusters of languages that are geographically close or linguistically related (e.g., Germanic or Romance languages), such that each cluster would cover at least one language of WMT. We list all the languages in the appendix.

\paragraph{Results} Table~\ref{table:distillation-results} presents performance results on WMT Metrics 2020. For each student model, we present the performance of a naive fine-tuning baseline, followed by vanilla pre-trained distillation on WMT data. We then introduce our synthetic data and 1-to-N distillation. We compare to \textsc{Comet}, \textsc{Prism}, and \BLEURText{}, three SOTA metrics from WMT Metrics '20~\cite{mathur2020results}.

On \emph{en-*}, the improvements are cumulative:  \emph{Distill WMT+Wiki} outperforms \emph{Distill WMT} (between 5 and 12\% improvement), and it is itself outperformed by \emph{1-to-N} (up to 4\%). Combining techniques improves the baselines in all cases, up to 10.5\% improvement compared to fine-tuning. RemBERT-12 matches 92.6\% of the teacher model's performance using only a third of its parameters, and it is competitive with current state-of-the-art models.

\begin{figure}[t!]
    \centering
    \includegraphics[width=\columnwidth]{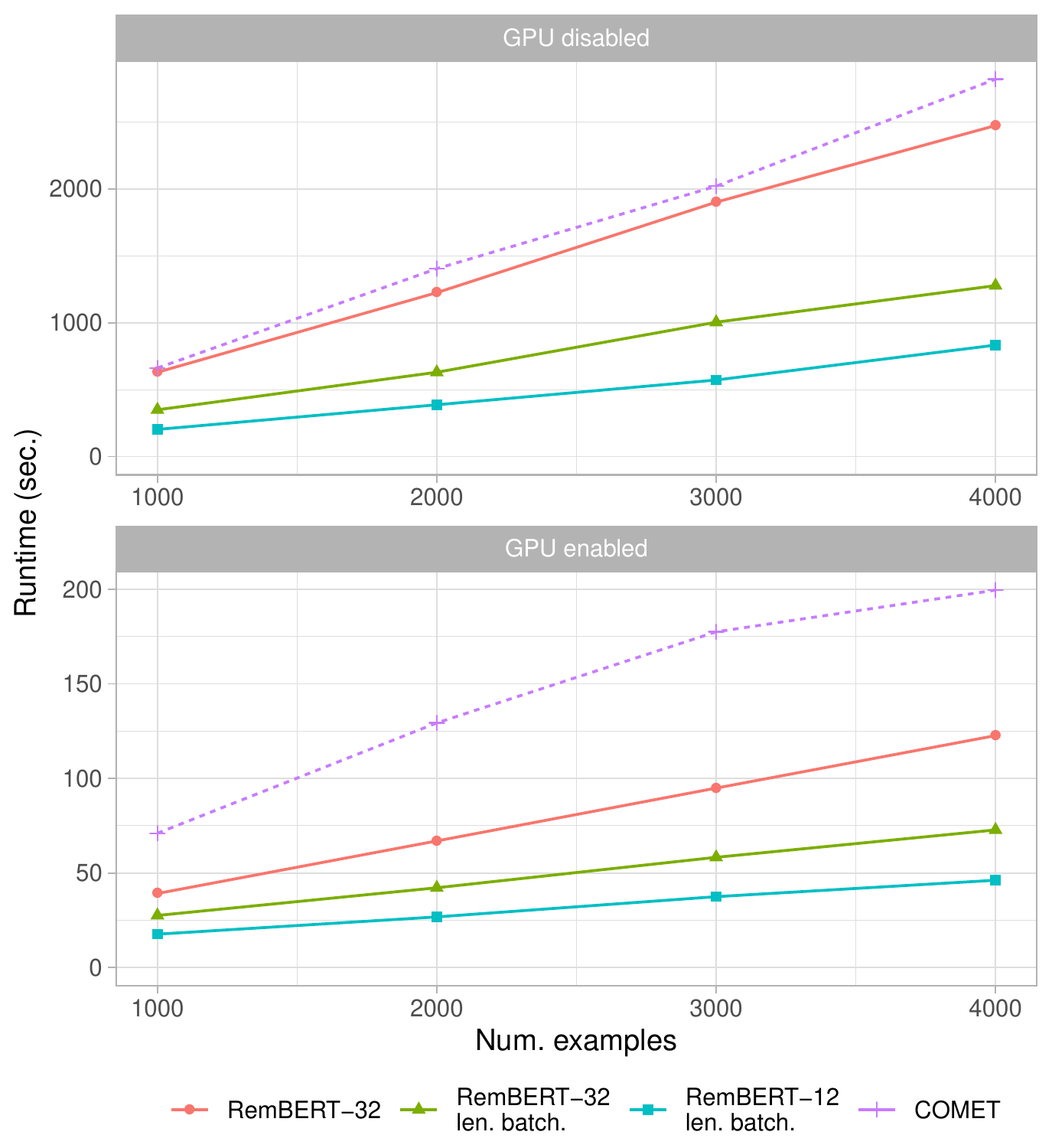}
    \caption{Runtime varying number of examples, with and without GPU. We ran each experiment 3 times and averaged the results, using a Google Cloud VM with 4 Intel Haswell vCPUS, 15GB main memory, and an nVidia Tesla T4 GPU. We used COMET 0.1.0, with the model \texttt{wmt-large-da-estimator-1719}.}
    \label{fig:runtime}
\end{figure}

\paragraph{Runtime} To validate the usefulness of our approach, we illustrate how to speed up RemBERT in Figure~\ref{fig:runtime}.
We obtain a first ~1.5-2X speedup compared to RemBERT-32 by applying length-based batching, a simple optimization which consists in batching examples that have similar a length and cropping the resulting tensor, as done in BERT-Score~\cite{zhang2019bertscore}. Doing so allows us to remove the padding tokens, which cause wasteful computations. We obtain a further ~1.5X speed-up by using the distilled version of the model, RemBERT-12. The final model processes 4.8 tuples per second without GPU (86 with a GPU), an 2.5-3X improvement over RemBERT-32.

Note that RemBERT-32 and COMET are both based on the Transformer architecture (we used a COMET checkpoint based on XLM-R Large), and RemBERT-32 is larger than COMET. We hypothesize that the performance gap comes from differences in implementation and model architecture; in particular, RemBERT-32 has an input sequence length of 128 while XLM-R operates on sequences with length 512.




\section{Conclusion}
We experimented with cross-lingual transfer in learned metrics, exposed the trade-off between multilinguality and model capacity, and addressed the problem with distillation on synthetic data. Further work includes generalizing the approach other tasks and experimenting with complementary compression methods such as pruning and quantization~\cite{kim2021bert, sanh2020movement}, as well as increasing linguistic coverage~\cite{joshi-etal-2020-state}.

\section*{Acknowledgments}
We thank Vitaly Nikolaev, who provided guidance on language families and created groups for the multiple-students setup. We also thank Iulia Turc, Shashi Narayan, George Foster, Markus Freitag, and Sebastian Ruder for the proof-reading, feedback, and suggestions.

\bibliography{main}

\begin{thebibliography}{36}
\expandafter\ifx\csname natexlab\endcsname\relax\def\natexlab#1{#1}\fi

\bibitem[{Aharoni et~al.(2019)Aharoni, Johnson, and
  Firat}]{aharoni2019massively}
Roee Aharoni, Melvin Johnson, and Orhan Firat. 2019.
\newblock \href {https://doi.org/10.18653/v1/N19-1388} {Massively multilingual
  neural machine translation}.
\newblock In \emph{Proceedings of the 2019 Conference of the North {A}merican
  Chapter of the Association for Computational Linguistics: Human Language
  Technologies, Volume 1 (Long and Short Papers)}, pages 3874--3884,
  Minneapolis, Minnesota. Association for Computational Linguistics.

\bibitem[{Belz and Reiter(2006)}]{belz2006comparing}
Anja Belz and Ehud Reiter. 2006.
\newblock \href {https://www.aclweb.org/anthology/E06-1040} {Comparing
  automatic and human evaluation of {NLG} systems}.
\newblock In \emph{11th Conference of the {E}uropean Chapter of the Association
  for Computational Linguistics}, Trento, Italy. Association for Computational
  Linguistics.

\bibitem[{Callison-Burch et~al.(2006)Callison-Burch, Osborne, and
  Koehn}]{callison2006re}
Chris Callison-Burch, Miles Osborne, and Philipp Koehn. 2006.
\newblock \href {https://www.aclweb.org/anthology/E06-1032} {Re-evaluating the
  role of {B}leu in machine translation research}.
\newblock In \emph{11th Conference of the {E}uropean Chapter of the Association
  for Computational Linguistics}, Trento, Italy. Association for Computational
  Linguistics.

\bibitem[{Celikyilmaz et~al.(2020)Celikyilmaz, Clark, and
  Gao}]{celikyilmaz2020evaluation}
Asli Celikyilmaz, Elizabeth Clark, and Jianfeng Gao. 2020.
\newblock \href {http://arxiv.org/abs/2006.14799} {Evaluation of text
  generation: {A} survey}.
\newblock \emph{CoRR}, abs/2006.14799.

\bibitem[{Chung et~al.(2021)Chung, Fevry, Tsai, Johnson, and
  Ruder}]{chung2020rethinking}
Hyung~Won Chung, Thibault Fevry, Henry Tsai, Melvin Johnson, and Sebastian
  Ruder. 2021.
\newblock \href {https://openreview.net/forum?id=xpFFI_NtgpW} {Rethinking
  embedding coupling in pre-trained language models}.
\newblock In \emph{International Conference on Learning Representations}.

\bibitem[{Conneau et~al.(2020{\natexlab{a}})Conneau, Khandelwal, Goyal,
  Chaudhary, Wenzek, Guzm{\'a}n, Grave, Ott, Zettlemoyer, and
  Stoyanov}]{conneau2019unsupervised}
Alexis Conneau, Kartikay Khandelwal, Naman Goyal, Vishrav Chaudhary, Guillaume
  Wenzek, Francisco Guzm{\'a}n, Edouard Grave, Myle Ott, Luke Zettlemoyer, and
  Veselin Stoyanov. 2020{\natexlab{a}}.
\newblock \href {https://doi.org/10.18653/v1/2020.acl-main.747} {Unsupervised
  cross-lingual representation learning at scale}.
\newblock In \emph{Proceedings of the 58th Annual Meeting of the Association
  for Computational Linguistics}, pages 8440--8451, Online. Association for
  Computational Linguistics.

\bibitem[{Conneau et~al.(2020{\natexlab{b}})Conneau, Khandelwal, Goyal,
  Chaudhary, Wenzek, Guzm{\'a}n, Grave, Ott, Zettlemoyer, and
  Stoyanov}]{conneau-etal-2020-unsupervised}
Alexis Conneau, Kartikay Khandelwal, Naman Goyal, Vishrav Chaudhary, Guillaume
  Wenzek, Francisco Guzm{\'a}n, Edouard Grave, Myle Ott, Luke Zettlemoyer, and
  Veselin Stoyanov. 2020{\natexlab{b}}.
\newblock \href {https://doi.org/10.18653/v1/2020.acl-main.747} {Unsupervised
  cross-lingual representation learning at scale}.
\newblock In \emph{Proceedings of the 58th Annual Meeting of the Association
  for Computational Linguistics}, pages 8440--8451, Online. Association for
  Computational Linguistics.

\bibitem[{Conneau and Lample(2019)}]{conneau2019cross}
Alexis Conneau and Guillaume Lample. 2019.
\newblock \href
  {https://proceedings.neurips.cc/paper/2019/file/c04c19c2c2474dbf5f7ac4372c5b9af1-Paper.pdf}
  {Cross-lingual language model pretraining}.
\newblock In \emph{Advances in Neural Information Processing Systems},
  volume~32.

\bibitem[{Conneau et~al.(2018)Conneau, Rinott, Lample, Williams, Bowman,
  Schwenk, and Stoyanov}]{conneau2018xnli}
Alexis Conneau, Ruty Rinott, Guillaume Lample, Adina Williams, Samuel Bowman,
  Holger Schwenk, and Veselin Stoyanov. 2018.
\newblock \href {https://doi.org/10.18653/v1/D18-1269} {{XNLI}: Evaluating
  cross-lingual sentence representations}.
\newblock In \emph{Proceedings of the 2018 Conference on Empirical Methods in
  Natural Language Processing}, pages 2475--2485, Brussels, Belgium.
  Association for Computational Linguistics.

\bibitem[{Devlin et~al.(2019)Devlin, Chang, Lee, and
  Toutanova}]{devlin2018bert}
Jacob Devlin, Ming-Wei Chang, Kenton Lee, and Kristina Toutanova. 2019.
\newblock \href {https://doi.org/10.18653/v1/N19-1423} {{BERT}: Pre-training of
  deep bidirectional transformers for language understanding}.
\newblock In \emph{Proceedings of the 2019 Conference of the North {A}merican
  Chapter of the Association for Computational Linguistics: Human Language
  Technologies, Volume 1 (Long and Short Papers)}, pages 4171--4186,
  Minneapolis, Minnesota. Association for Computational Linguistics.

\bibitem[{Freitag et~al.(2021)Freitag, Foster, Grangier, Ratnakar, Tan, and
  Macherey}]{freitag2021experts}
Markus Freitag, George~F. Foster, David Grangier, Viresh Ratnakar, Qijun Tan,
  and Wolfgang Macherey. 2021.
\newblock \href {http://arxiv.org/abs/2104.14478} {Experts, errors, and
  context: {A} large-scale study of human evaluation for machine translation}.
\newblock \emph{CoRR}, abs/2104.14478.

\bibitem[{Hinton et~al.(2015)Hinton, Vinyals, and Dean}]{hinton2015distilling}
Geoffrey Hinton, Oriol Vinyals, and Jeffrey Dean. 2015.
\newblock \href {http://arxiv.org/abs/1503.02531} {Distilling the knowledge in
  a neural network}.
\newblock In \emph{NIPS Deep Learning and Representation Learning Workshop}.

\bibitem[{Joshi et~al.(2020)Joshi, Santy, Budhiraja, Bali, and
  Choudhury}]{joshi-etal-2020-state}
Pratik Joshi, Sebastin Santy, Amar Budhiraja, Kalika Bali, and Monojit
  Choudhury. 2020.
\newblock \href {https://doi.org/10.18653/v1/2020.acl-main.560} {The state and
  fate of linguistic diversity and inclusion in the {NLP} world}.
\newblock In \emph{Proceedings of the 58th Annual Meeting of the Association
  for Computational Linguistics}, pages 6282--6293, Online. Association for
  Computational Linguistics.

\bibitem[{Kim et~al.(2021)Kim, Gholami, Yao, Mahoney, and
  Keutzer}]{kim2021bert}
Sehoon Kim, Amir Gholami, Zhewei Yao, Michael~W Mahoney, and Kurt Keutzer.
  2021.
\newblock I-bert: Integer-only bert quantization.
\newblock \emph{arXiv preprint arXiv:2101.01321}.

\bibitem[{Kingma and Ba(2015)}]{Kingma2015}
Diederik~P. Kingma and Jimmy Ba. 2015.
\newblock \href {http://arxiv.org/abs/1412.6980} {Adam: {A} method for
  stochastic optimization}.
\newblock In \emph{3rd International Conference on Learning Representations,
  {ICLR} 2015, San Diego, CA, USA, May 7-9, 2015, Conference Track
  Proceedings}.

\bibitem[{Lin(2004)}]{lin2004rouge}
Chin-Yew Lin. 2004.
\newblock \href {https://www.aclweb.org/anthology/W04-1013} {{ROUGE}: A package
  for automatic evaluation of summaries}.
\newblock In \emph{Text Summarization Branches Out}, pages 74--81, Barcelona,
  Spain. Association for Computational Linguistics.

\bibitem[{Ma et~al.(2018)Ma, Bojar, and Graham}]{ma2018results}
Qingsong Ma, Ond{\v{r}}ej Bojar, and Yvette Graham. 2018.
\newblock \href {https://doi.org/10.18653/v1/W18-6450} {Results of the {WMT}18
  metrics shared task: Both characters and embeddings achieve good
  performance}.
\newblock In \emph{Proceedings of the Third Conference on Machine Translation:
  Shared Task Papers}, pages 671--688, Belgium, Brussels. Association for
  Computational Linguistics.

\bibitem[{Ma et~al.(2019)Ma, Wei, Bojar, and Graham}]{ma2019results}
Qingsong Ma, Johnny Wei, Ond{\v{r}}ej Bojar, and Yvette Graham. 2019.
\newblock \href {https://doi.org/10.18653/v1/W19-5302} {Results of the {WMT}19
  metrics shared task: Segment-level and strong {MT} systems pose big
  challenges}.
\newblock In \emph{Proceedings of the Fourth Conference on Machine Translation
  (Volume 2: Shared Task Papers, Day 1)}, pages 62--90, Florence, Italy.
  Association for Computational Linguistics.

\bibitem[{Mathur et~al.(2020{\natexlab{a}})Mathur, Baldwin, and
  Cohn}]{mathur2020tangled}
Nitika Mathur, Timothy Baldwin, and Trevor Cohn. 2020{\natexlab{a}}.
\newblock \href {https://doi.org/10.18653/v1/2020.acl-main.448} {Tangled up in
  {BLEU}: Reevaluating the evaluation of automatic machine translation
  evaluation metrics}.
\newblock In \emph{Proceedings of the 58th Annual Meeting of the Association
  for Computational Linguistics}, pages 4984--4997, Online. Association for
  Computational Linguistics.

\bibitem[{Mathur et~al.(2020{\natexlab{b}})Mathur, Wei, Freitag, Ma, and
  Bojar}]{mathur2020results}
Nitika Mathur, Johnny Wei, Markus Freitag, Qingsong Ma, and Ond{\v{r}}ej Bojar.
  2020{\natexlab{b}}.
\newblock \href {https://www.aclweb.org/anthology/2020.wmt-1.77} {Results of
  the {WMT}20 metrics shared task}.
\newblock In \emph{Proceedings of the Fifth Conference on Machine Translation},
  pages 688--725, Online. Association for Computational Linguistics.

\bibitem[{Papineni et~al.(2002)Papineni, Roukos, Ward, and
  Zhu}]{papineni2002bleu}
Kishore Papineni, Salim Roukos, Todd Ward, and Wei-Jing Zhu. 2002.
\newblock \href {https://doi.org/10.3115/1073083.1073135} {{BLEU}: a method for
  automatic evaluation of machine translation}.
\newblock In \emph{Proceedings of the 40th Annual Meeting of the Association
  for Computational Linguistics}, pages 311--318, Philadelphia, Pennsylvania,
  USA. Association for Computational Linguistics.

\bibitem[{Peters et~al.(2018)Peters, Neumann, Iyyer, Gardner, Clark, Lee, and
  Zettlemoyer}]{peters2018deep}
Matthew Peters, Mark Neumann, Mohit Iyyer, Matt Gardner, Christopher Clark,
  Kenton Lee, and Luke Zettlemoyer. 2018.
\newblock \href {https://doi.org/10.18653/v1/N18-1202} {Deep contextualized
  word representations}.
\newblock In \emph{Proceedings of the 2018 Conference of the North {A}merican
  Chapter of the Association for Computational Linguistics: Human Language
  Technologies, Volume 1 (Long Papers)}, pages 2227--2237, New Orleans,
  Louisiana. Association for Computational Linguistics.

\bibitem[{Pires et~al.(2019)Pires, Schlinger, and
  Garrette}]{pires2019multilingual}
Telmo Pires, Eva Schlinger, and Dan Garrette. 2019.
\newblock \href {https://doi.org/10.18653/v1/P19-1493} {How multilingual is
  multilingual {BERT}?}
\newblock In \emph{Proceedings of the 57th Annual Meeting of the Association
  for Computational Linguistics}, pages 4996--5001, Florence, Italy.
  Association for Computational Linguistics.

\bibitem[{Raffel et~al.(2020)Raffel, Shazeer, Roberts, Lee, Narang, Matena,
  Zhou, Li, and Liu}]{DBLP:journals/jmlr/RaffelSRLNMZLL20}
Colin Raffel, Noam Shazeer, Adam Roberts, Katherine Lee, Sharan Narang, Michael
  Matena, Yanqi Zhou, Wei Li, and Peter~J. Liu. 2020.
\newblock \href {http://jmlr.org/papers/v21/20-074.html} {Exploring the limits
  of transfer learning with a unified text-to-text transformer}.
\newblock \emph{J. Mach. Learn. Res.}, 21:140:1--140:67.

\bibitem[{Rei et~al.(2020{\natexlab{a}})Rei, Stewart, Farinha, and
  Lavie}]{rei2020comet}
Ricardo Rei, Craig Stewart, Ana~C Farinha, and Alon Lavie. 2020{\natexlab{a}}.
\newblock \href {https://doi.org/10.18653/v1/2020.emnlp-main.213} {{COMET}: A
  neural framework for {MT} evaluation}.
\newblock In \emph{Proceedings of the 2020 Conference on Empirical Methods in
  Natural Language Processing (EMNLP)}, pages 2685--2702, Online. Association
  for Computational Linguistics.

\bibitem[{Rei et~al.(2020{\natexlab{b}})Rei, Stewart, Farinha, and
  Lavie}]{rei-etal-2020-unbabels}
Ricardo Rei, Craig Stewart, Ana~C Farinha, and Alon Lavie. 2020{\natexlab{b}}.
\newblock \href {https://aclanthology.org/2020.wmt-1.101} {Unbabel{'}s
  participation in the {WMT}20 metrics shared task}.
\newblock In \emph{Proceedings of the Fifth Conference on Machine Translation},
  pages 911--920, Online. Association for Computational Linguistics.

\bibitem[{Sanh et~al.(2019)Sanh, Debut, Chaumond, and
  Wolf}]{sanh2019distilbert}
Victor Sanh, Lysandre Debut, Julien Chaumond, and Thomas Wolf. 2019.
\newblock Distilbert, a distilled version of bert: smaller, faster, cheaper and
  lighter.
\newblock In \emph{NeurIPS 5th Workshop on Energy Efficient Machine Learning
  and Cognitive Computing}.

\bibitem[{Sanh et~al.(2020)Sanh, Wolf, and Rush}]{sanh2020movement}
Victor Sanh, Thomas Wolf, and Alexander Rush. 2020.
\newblock Movement pruning: Adaptive sparsity by fine-tuning.
\newblock In \emph{Advances in Neural Information Processing Systems},
  volume~33, pages 20378--20389. Curran Associates, Inc.

\bibitem[{Sellam et~al.(2020{\natexlab{a}})Sellam, Das, and
  Parikh}]{sellam2020bleurt}
Thibault Sellam, Dipanjan Das, and Ankur Parikh. 2020{\natexlab{a}}.
\newblock \href {https://doi.org/10.18653/v1/2020.acl-main.704} {{BLEURT}:
  Learning robust metrics for text generation}.
\newblock In \emph{Proceedings of the 58th Annual Meeting of the Association
  for Computational Linguistics}, pages 7881--7892, Online. Association for
  Computational Linguistics.

\bibitem[{Sellam et~al.(2020{\natexlab{b}})Sellam, Pu, Chung, Gehrmann, Tan,
  Freitag, Das, and Parikh}]{sellam2020learning}
Thibault Sellam, Amy Pu, Hyung~Won Chung, Sebastian Gehrmann, Qijun Tan, Markus
  Freitag, Dipanjan Das, and Ankur Parikh. 2020{\natexlab{b}}.
\newblock \href {https://www.aclweb.org/anthology/2020.wmt-1.102} {Learning to
  evaluate translation beyond {E}nglish: {BLEURT} submissions to the {WMT}
  metrics 2020 shared task}.
\newblock In \emph{Proceedings of the Fifth Conference on Machine Translation},
  pages 921--927, Online. Association for Computational Linguistics.

\bibitem[{Shu et~al.(2021)Shu, Yoo, and Ha}]{shu2021reward}
Raphael Shu, Kang~Min Yoo, and Jung{-}Woo Ha. 2021.
\newblock \href {http://arxiv.org/abs/2104.07541} {Reward optimization for
  neural machine translation with learned metrics}.
\newblock \emph{CoRR}, abs/2104.07541.

\bibitem[{Thompson and Post(2020)}]{thompson-post-2020-automatic}
Brian Thompson and Matt Post. 2020.
\newblock \href {https://doi.org/10.18653/v1/2020.emnlp-main.8} {Automatic
  machine translation evaluation in many languages via zero-shot paraphrasing}.
\newblock In \emph{Proceedings of the 2020 Conference on Empirical Methods in
  Natural Language Processing (EMNLP)}, pages 90--121, Online. Association for
  Computational Linguistics.

\bibitem[{Turc et~al.(2019)Turc, Chang, Lee, and Toutanova}]{turc2019well}
Iulia Turc, Ming-Wei Chang, Kenton Lee, and Kristina Toutanova. 2019.
\newblock Well-read students learn better: The impact of student initialization
  on knowledge distillation.
\newblock \emph{arXiv}.

\bibitem[{{Xue} et~al.(2020){Xue}, {Constant}, {Roberts}, {Kale}, {Al-Rfou},
  {Siddhant}, {Barua}, and {Raffel}}]{xue_mt5}
Linting {Xue}, Noah {Constant}, Adam {Roberts}, Mihir {Kale}, Rami {Al-Rfou},
  Aditya {Siddhant}, Aditya {Barua}, and Colin {Raffel}. 2020.
\newblock \href {http://arxiv.org/abs/2010.11934} {{mT5: A massively
  multilingual pre-trained text-to-text transformer}}.
\newblock \emph{arXiv e-prints}, page arXiv:2010.11934.

\bibitem[{Yang et~al.(2019)Yang, Dai, Yang, Carbonell, Salakhutdinov, and
  Le}]{yang2019xlnet}
Zhilin Yang, Zihang Dai, Yiming Yang, Jaime~G. Carbonell, Ruslan Salakhutdinov,
  and Quoc~V. Le. 2019.
\newblock \href
  {https://proceedings.neurips.cc/paper/2019/hash/dc6a7e655d7e5840e66733e9ee67cc69-Abstract.html}
  {Xlnet: Generalized autoregressive pretraining for language understanding}.
\newblock In \emph{Advances in Neural Information Processing Systems 32: Annual
  Conference on Neural Information Processing Systems 2019, NeurIPS 2019,
  December 8-14, 2019, Vancouver, BC, Canada}, pages 5754--5764.

\bibitem[{Zhang* et~al.(2020)Zhang*, Kishore*, Wu*, Weinberger, and
  Artzi}]{zhang2019bertscore}
Tianyi Zhang*, Varsha Kishore*, Felix Wu*, Kilian~Q. Weinberger, and Yoav
  Artzi. 2020.
\newblock \href {https://openreview.net/forum?id=SkeHuCVFDr} {Bertscore:
  Evaluating text generation with bert}.
\newblock In \emph{International Conference on Learning Representations}.

\end{thebibliography}
\bibliographystyle{acl_natbib}

\appendix

\section{Training RemBERT for MT Evaluation}
\label{appendix:training}

\subsection{RemBERT Pre-Training}
\label{apendix:rembert}

\begin{table*}[!ht]
\small
\begin{center}
\begin{tabular}{lcccc}
\toprule
Hyperparameter & RemBERT & RemBERT-3 & RemBERT-6 & RemBERT-12 \\
\midrule
Number of layers & 32 & 3 &6 & 12\\
Hidden size & 1152 & 640 & 640 & 1024\\
Vocabulary size & 250,000 & 120,000 & 120,000 & 120,000 \\
Input embedding dimension & 256 & 128 & 128 & 128 \\
Output embedding dimension & 1536 & 2048 & 2048 & 2048\\
Number of heads & 18 & 8 & 8 & 16\\
Head dimension & 64 & 80 & 80 & 64\\
\midrule
Num. params. during pre-training  & 995M & 276M & 291M & 412M \\
Num. params. during fine-tuning & 579M & 30M & 45M & 167M \\
\bottomrule
\end{tabular}
\end{center}
\caption{RemBERT architecture.}
\label{table:remberthparams}
\end{table*}

RemBERT is an encoder-only architecture, similar to BERT but with an optimized parameter allocation~\cite{chung2020rethinking}. It has reduced input embedding dimension and the saved parameters are reinvested in the form of wider and deeper Transformer layers, keeping the model size constant. In addition, the input and the output embeddings (the weights associated with the softmax layer) are decoupled during pre-training.

Table~\ref{table:remberthparams} describes the architecture of the four RemBERT models, along with the number of parameters (note that we remove the output embedding layer during fine-tuning, which reduces the model size). We obtained the original RemBERT model from its authors, and we trained the smaller models for the purpose of this study with a modified version of the public BERT codebase.\footnote{\url{https://github.com/google-research/bert}} By default, all models are pre-trained on 104 languages using a masked language modelling objective~\cite{devlin2018bert}. The setup for the smaller models is similar to \citet{chung2020rethinking}, except that RemBERT uses on mC4~\citep{xue_mt5} and Wikipedia while we use Wikipedia only. We train the custom RemBERT models for $2^{17}$ steps using the Adam optimizer~\citep{Kingma2015}, using learning rate 0.0002 (with 10,000 linear warm-up followed by inverse square root decay schedule) and batch size 512 on 16 TPU v3 chips. To reduce the size of the models further, we use a smaller SentencePiece model with 120K tokens instead of 250k. Large RemBERT was fine-tuned with sequence size 128, while the student models were fine-tuned with sequence size 512.

\subsection{Fine-Tuning for the WMT Metrics Shared Task}
\label{sec:finetuning}

\begin{figure}
    \centering
    \includegraphics[width=\columnwidth]{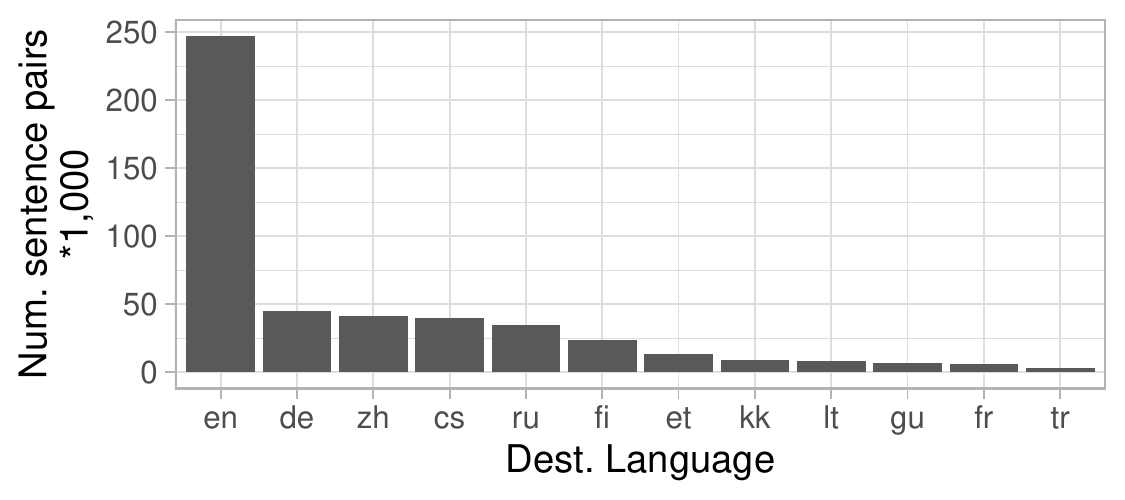}
    \caption{Distribution of the languages in the training set, built from WMT Metrics 2015 to 2019.}
    \label{fig:wmt-langs}
\end{figure}

We fine-tune RemBERT on the WMT Metrics shared task following the methology of~\citet{sellam2020learning}.  
We combine all the sentence pairs of WMT 2015 to 2019, and set aside 5\% of the data for continuous evaluation. The data can be downloaded from the WMT Website.\footnote{\url{http://www.statmt.org/wmt20/metrics-task.html}} The distribution of examples per language is shown in Figure.~\ref{fig:wmt-langs}. We sample the sentences randomly, then re-adjust the sample such that there are not reference translations leaking between the datasets. We train the model with Adam for 5,000 steps and a batch size of 128 while evaluating it on the eval set every 250 steps. We keep the checkpoint that leads to the best performance. To determine the learning rate, we ran a parameter sweep on a previous year of the benchmark (using 2015 to 18 for train and 2019 for test) using the values \texttt{[1e-6, 2e-6, 5e-6, 7e-6, 8e-6, 9e-6, 1e-5, 2e-5]}, and kept the learning rate that led the best results (\texttt{1e-6}). We also experimented with language rebalancing, batch sizes, dropout, and training duration during preliminary sets of experiments. The setup we used for RemBERT-3, 6 and 12 is similar, except that we used learning rate 1e-5 (obtained with a parameter sweep on a randomly held-out sample), 20,000 training steps, batch size 32, and we evaluate the model every 1,000 steps. We train each model with 4 TPU v2 chips, and evaluate with a Nvidia Tesla V100 GPU.

\section{Additional Ablation Experiments on WMT Metrics Shared task 2020}
\label{sec:ablations}

\begin{figure*}[!ht]
    \centering
    \includegraphics[width=\textwidth]{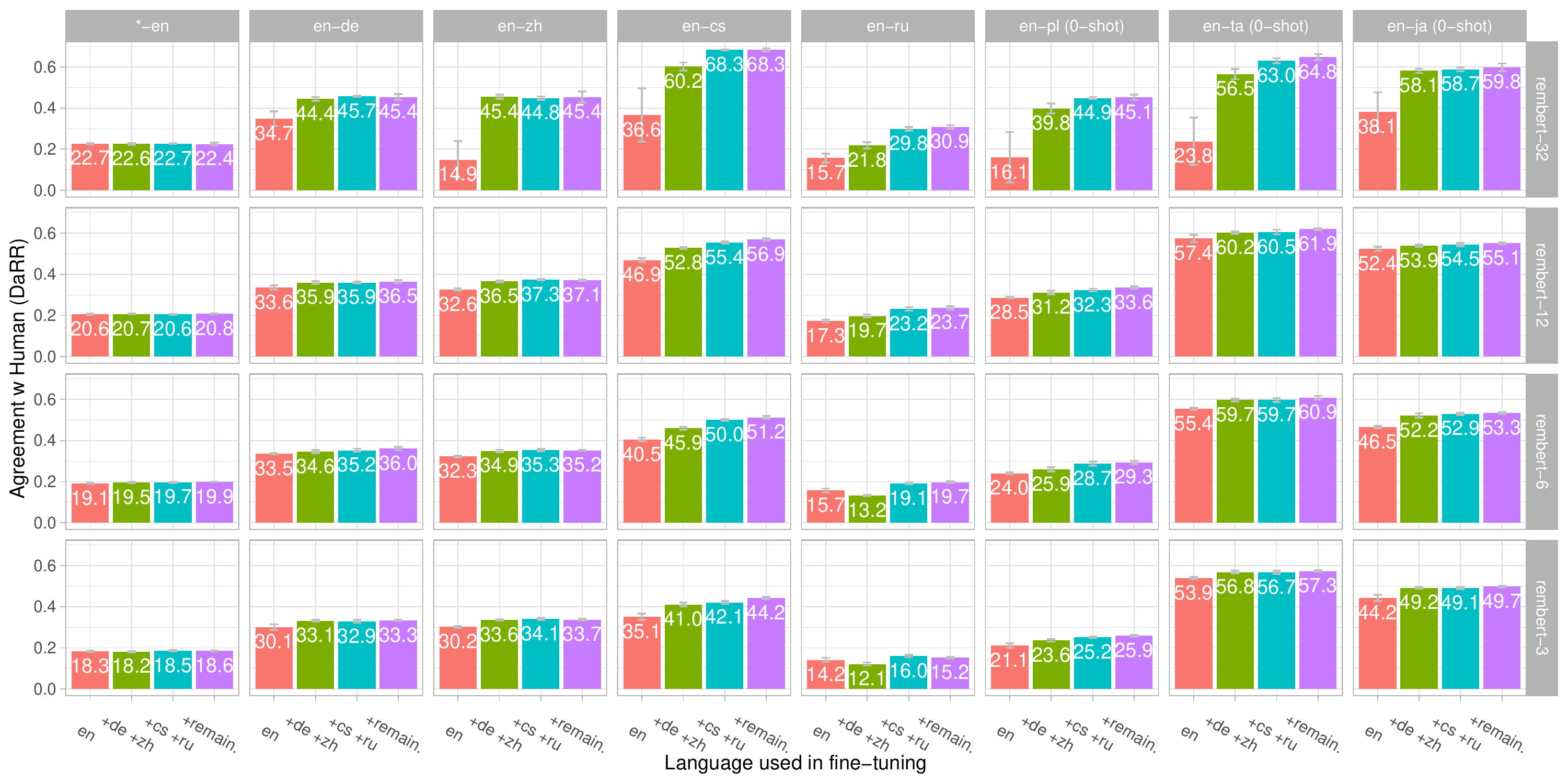}
    \caption{Impact of the number of fine-tuning languages on BLEURT's performance, using all models and all test language pairs of WMT'20 except \texttt{en-iu}. The metric is WMT Metrics DaRR~\cite{mathur2020results}, a robust variant of Kendall Tau, higher is better. We run each experiment 5 times, report the mean result and Normal-based 95\% confidence intervals.}
    \label{fig:ft-langs-all}
\end{figure*}

\begin{figure*}[!ht]
    \centering
    \includegraphics[width=\textwidth]{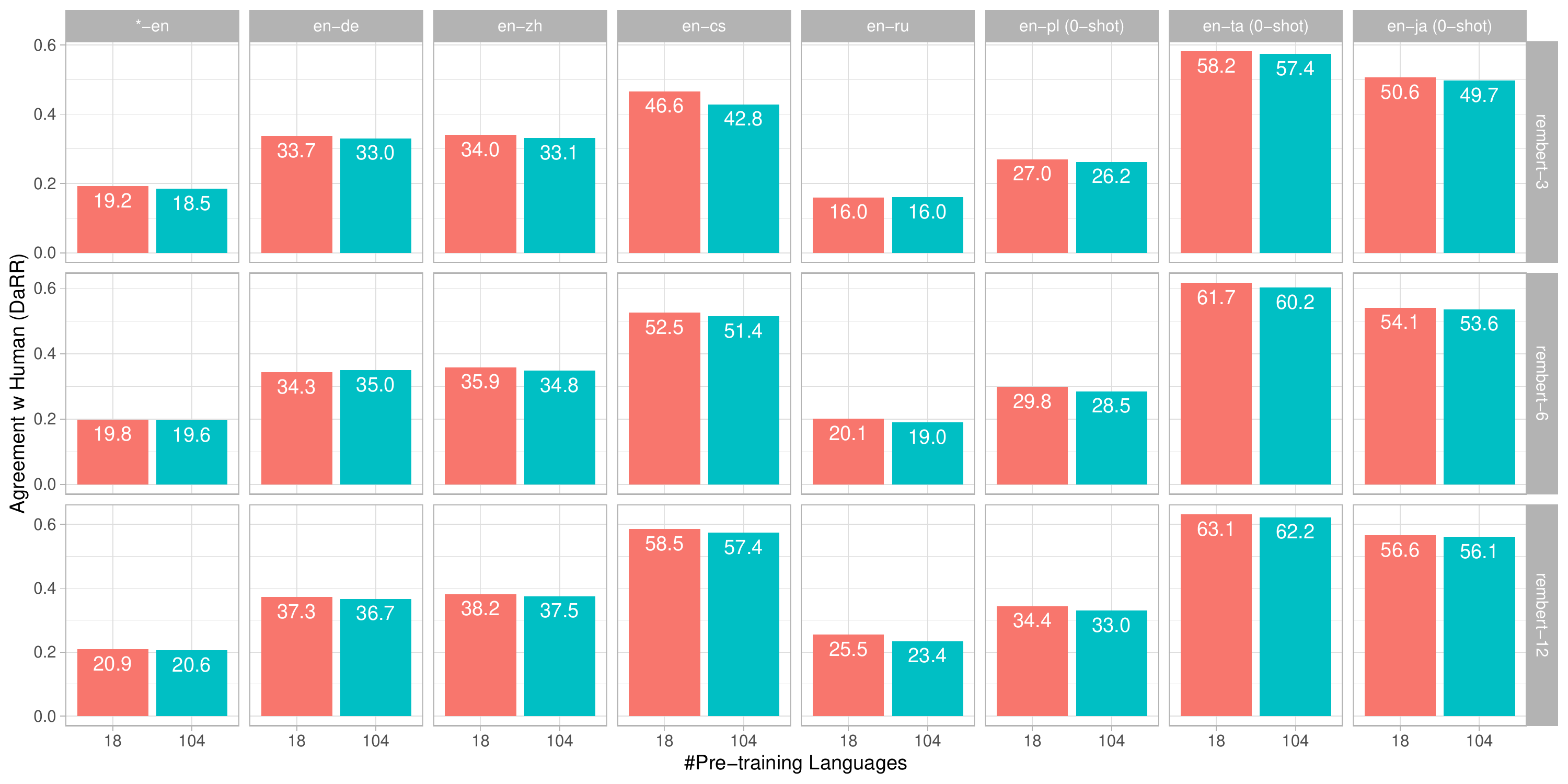}
    \caption{Performance improvement after removing 86 languages from from pre-training (out of 104), using all models and all test language pairs of WMT'20 except \texttt{en-iu}.}
    \label{fig:pt-langs-all}
\end{figure*}

We present the detail of our ablation experiments, which expose the trade-off between model capacity and multi-linguality in learned metrics.

In Figure~\ref{fig:ft-langs-all}, we iteratively expand the number of fine-tuning languages, starting with only English and adding languages in decreasing order of frequency. We add the languages by bucket, such that each bucket contains about the same number of examples (Figure~\ref{fig:wmt-langs} shows the size of the training set for each language). 

We start with the five languages for which we have training data. In all cases introducing fine-tuning data for a particular language pair improves the metric's performance on this language. The effect of subsequent additions (that is, cross-lingual transfer) is mixed. For instance, the effect is mild to negative on \texttt{*-en}, while it is mostly positive \texttt{en-cs}. 

Adding data has a different effect on zero-shot languages: in almost all cases, it brings improvements. The effect appears milder on the smaller models, especially RemBERT-3 for which we observe slight performance drops (\texttt{en-ta} and \texttt{en-ja}), which is consistent the ``curse of multilinguality''~\cite{conneau2019cross}.

Figure~\ref{fig:pt-langs-all} shows the limit of our smaller models: in 21 cases out of 24 (regardless of whether the language is zero-shot or not), the performance of the model improves when we remove 86 languages from pre-training. This is further evidence that the models are saturated.

\section{Details of the Distillation Pipeline}
\begin{table*}[ht!]
\small
\centering
\begin{tabular}{@{}ccccc@{}}
\toprule
Student 1 & Student 2 &  Student 3 &  Student 4 &  Student 5 \\ \midrule
\makecell[tc]{
Afrikaans\\
Danish\\
Dutch\\
English\\
German\\
Icelandic\\
Luxembourgish\\
Norwegian\\
Swedish\\
West Frisian} & 
\makecell[tc]{
Catalan \\
French \\
Galician \\
Haitian Creole \\
Italian \\
Latin \\
Portuguese \\
Romanian \\
Spanish}& 
\makecell[tc]{
Bengali \\
Gujarati \\
Hindi \\
Hindi (Latin) \\
Marathi \\
Nepali \\
Persian \\
Punjabi \\
Tajik \\
Urdu \\
Tamil} & 
\makecell[tc]{
Belarusian \\
Bulgarian \\
Bulgarian (Latin) \\
Czech \\
Macedonian \\
Polish \\
Russian \\
Russian (Latin) \\
Serbian \\
Slovak \\
Slovenian \\
Ukranian \\
Finnish \\
Estonian \\
Kazakh \\
Lithuanian \\
Latvian \\
Turkish} & 
\makecell[tc]{
Burmese \\
Chinese \\
Chinese (Latin) \\
Japanese} \\ \bottomrule
\end{tabular}
\caption{Languages per student used for 1-to-N distillation. }
\label{tab:language-clusters}
\end{table*}

\subsection{Distillation Data Generation Method}
\label{sec:perturbations}

We generate synthetic \texttt{(Reference Translation, MT outputs)} pairs by perturbing sentences from Wikipedia. A similar method has been shown to be useful when generating pre-training data in a monolingual context~\cite{sellam2020bleurt}. We apply it to 13 languages: English, Japanese, Lithuanian, Czech, Tamil, Chinese, Russian, Kazakh, Gujarati, Finnish, French, Polish, and German. We chose these languages because they are covered by the WMT Metrics setup during training (e.g., Kazakh, Gujarati, Finnish), testing (Japanese, Polish, Tamil), or both. We emulate the noise introduced by MT systems with three types of perturbations:
\begin{itemize}
\item Word substitution: we randomly mask up to 15 WordPiece tokens, and replace the masks by a multilingual model. We sample the number of tokens to be masked uniformly, and we run beam search with mBERT, using beam size 8. We used the official mBERT model.\footnote{\url{https://github.com/google-research/bert}}
\item Back-translation: we translate the Wikipedia from the source to English, then back in the source language with translation models. We used the Tensor2Tensor framework,\footnote{\url{https://github.com/tensorflow/tensor2tensor}} using models trained on the corresponding WMT datasets.
\item Word dropping: we duplicate 30\% of the dataset and randomly drop words from the perturbations.
\end{itemize}

We generate between 1.8M and 7.3M sentence pairs for each language, for a total of 84M un-labelled examples.

\subsection{Languages Used in 1-to-N Distillation}
\label{sec:languages-appendix}

Table~\ref{tab:language-clusters} shows the five language clusters used for the 1-to-N distillation experiments. The groups were created by first joining languages based on their linguistic proximity (e.g., Romance or Germanic languages). Since that left multiple languages in their own cluster, we then combined them based on geographic distance (e.g., Tamil is part of the otherwise Indo-Iranian cluster and Japanese part of a cluster of Sino-Tibetan languages).  

\subsection{Setup and Hyper-parameters}

The hyper-parameters we use for distillation are similar to those of fine-tuning, except that we train the models for 500,000 batches of 128 examples, and thus we learn from 64M sentences instead of 640K. Doing so takes about 1.5 days for RemBERT-3 and 6, and 3.5 days for RemBERT-12. We train the models to completion (i.e., no early-stopping).

\section{Additional Details of Metrics Performance}
\label{sec:performance}

\begin{table*}[t]
\centering
\scriptsize
\begin{tabular} { l l | c | c c c c c c c c c c }
\toprule
& & *-en & cs-en & de-en & iu-en$^\dagger$ & ja-en & km-en$^\dagger$ & pl-en & ps-en$^\dagger$ & ru-en & ta-en & zh-en\\
\midrule
\multicolumn{2}{l |}{BLEURT-Tiny~\cite{sellam2020bleurt}} & 17.0 & 5.2 & 42.1 & 22.7 & 18.4 & 26.5 & 2.1 & 16.8 & 3.8 & 21.7 & 10.7\\
\multicolumn{2}{l |}{BLEURT~\cite{sellam2020bleurt}} & 21.2 & 10.7 & 45.4 & 27.4 & 25.8 & 31.7 & 4.1 & 21.1 & 8.0 & 24.0 & 13.9\\
\multicolumn{2}{l |}{BLEURT English WMT'20~\cite{sellam2020learning}} & 22.1 & 12.6 & 45.3 & 27.6 & 26.5 & 33.3 & 5.7 & 23.5 & 9.3 & 23.1 & 13.7\\
\multicolumn{2}{l |}{BLEURT-Ext~\cite{sellam2020learning}} & 22.0 & 12.7 & 44.6 & 27.9 & 27.1 & 33.8 & 4.4 & 20.8 & 10.1 & 24.7 & 13.7\\
\midrule
RemBERT-32 & &  22.5 & 13.9 & 46.2 & 29.1 & 28.0 & 31.1 & 4.6 & 22.6 & 9.9 & 25.1 & 14.5\\
\midrule
rembert-3 & Fine-tuning & 18.5 & 8.0 & 42.7 & 25.0 & 23.3 & 26.9 & 2.6 & 19.1 & 4.9 & 19.7 & 12.8\\
 & N Fine-tuning & 18.0 & 8.1 & 42.8 & 24.1 & 23.0 & 26.5 & 2.8 & 17.4 & 4.6 & 19.1 & 11.8\\
 & Distill. WMT & 16.3 & 7.0 & 41.2 & 23.6 & 21.5 & 24.1 & 0.1 & 12.6 & 2.8 & 19.3 & 11.2\\
 & Distill. Wiki + WMT & 19.1 & 7.5 & 43.6 & 24.7 & 25.1 & 25.7 & 2.8 & 19.2 & 7.9 & 21.5 & 13.3\\
 & 1-to-N Distill & 19.9 & 9.1 & 44.8 & 27.0 & 24.4 & 27.9 & 1.5 & 20.4 & 7.4 & 22.7 & 13.7\\
 \midrule
rembert-6 & Fine-tuning & 19.6 & 10.1 & 43.9 & 24.6 & 23.4 & 29.5 & 4.0 & 20.2 & 6.5 & 20.4 & 13.4\\
 & N Fine-tuning & 18.9 & 9.5 & 44.2 & 22.9 & 21.6 & 28.9 & 1.9 & 20.4 & 6.4 & 21.3 & 12.3\\
 & Distill. WMT & 19.9 & 9.4 & 43.8 & 26.9 & 24.3 & 30.4 & 4.3 & 19.5 & 5.9 & 21.1 & 13.5\\
 & Distill. Wiki + WMT & 20.7 & 10.4 & 45.2 & 26.9 & 25.0 & 30.3 & 2.3 & 21.2 & 8.4 & 23.4 & 14.1\\
 & 1-to-N Distill & 21.0 & 10.8 & 45.7 & 26.1 & 25.4 & 29.5 & 4.0 & 21.5 & 9.2 & 23.8 & 14.0\\
\midrule
rembert-12 & Fine-tuning & 20.6 & 10.5 & 45.2 & 25.4 & 25.2 & 30.6 & 4.4 & 20.4 & 8.5 & 21.8 & 14.0\\
 & N Fine-tuning & 19.5 & 12.1 & 44.8 & 23.4 & 24.7 & 28.6 & 2.1 & 19.1 & 7.4 & 20.5 & 12.7\\
 & Distill. WMT & 21.4 & 11.1 & 45.6 & 25.4 & 26.7 & 31.1 & 6.0 & 21.7 & 8.6 & 23.6 & 14.5\\
 & Distill. Wiki + WMT & 21.9 & 11.9 & 45.8 & 28.8 & 26.0 & 31.6 & 4.6 & 22.6 & 10.1 & 23.5 & 14.1\\
 & 1-to-N Distill & 21.7 & 12.3 & 46.5 & 29.5 & 25.7 & 31.4 & 3.2 & 21.7 & 8.6 & 23.4 & 14.9\\
\bottomrule
\end{tabular}
\caption{Segment-level agreement with human ratings on to-English language pairs. The metric is WMT Metrics DaRR \cite{mathur2020results}, a robust variant of Kendall Tau, higher is better. The dagger$^\dagger{}$ indicates that our results may differ from at least 0.5 percentage point from the published WMT results on the language pair (we used the BLEURT submissions for comparison).}
\label{table:distillation-results-ext}
\end{table*}

\begin{table*}[!t]
\centering
\scriptsize
\begin{tabular} { l l | c | c c c c c c c }
\toprule
&  & en-* & en-cs & en-de & en-ja & en-pl & en-ru & en-ta & en-zh\\
\midrule
\multicolumn{2}{l |}{BLEURT-Ext.~\cite{sellam2020learning}} & 49.8 & 68.8 & 44.7 & 53.3 & 43.0 & 30.6 & 64.3 & 44.2\\
\multicolumn{2}{l |}{COMET~\cite{rei2020comet}$^\dagger{}$} &  52.4 & 66.8 & 46.8 & 62.4 & 46.2 & 34.4 & 67.1 & 43.2 \\
\multicolumn{2}{l |}{PRISM~\cite{thompson-post-2020-automatic}$^\dagger{}$} & 45.5 & 61.9 & 44.7 & 57.9 &41.4 & 28.3 &44.8 & 39.7 \\
\multicolumn{2}{l |}{YiSi-1~\cite{thompson-post-2020-automatic}$^\dagger{}$} & 46.9 &  55.0& 42.7& 56.8& 34.9 & 25.6 & 66.9 & 46.3\\
\midrule
RemBERT-32 & & 52.3 & 69.3 & 45.9 & 61.7 & 45.4 & 31.0 & 66.6 & 45.9\\
\midrule
rembert-3 & Fine-tuning & 36.9 & 42.8 & 33.0 & 49.7 & 26.2 & 16.0 & 57.4 & 33.1\\
 & N Fine-tuning & 32.7 & 42.7 & 30.8 & 43.3 & 21.8 & 13.7 & 44.6 & 31.7\\
 & Distill. WMT & 34.8 & 43.3 & 29.0 & 46.8 & 22.0 & 15.4 & 56.1 & 31.3\\
 & Distill. Wiki + WMT & 39.1 & 42.3 & 34.4 & 53.6 & 26.9 & 18.9 & 60.3 & 37.6\\
 & 1-to-N Distill & 40.1 & 47.3 & 32.9 & 54.4 & 27.3 & 19.3 & 60.0 & 39.6\\
\midrule
rembert-6 & Fine-tuning & 40.3 & 51.4 & 35.0 & 53.6 & 28.5 & 19.0 & 60.2 & 34.8\\
 & N Fine-tuning & 36.6 & 49.3 & 31.5 & 46.5 & 26.3 & 18.2 & 48.8 & 35.5\\
 & Distill. WMT & 40.4 & 53.1 & 34.8 & 52.1 & 28.4 & 17.9 & 60.1 & 36.3\\
 & Distill. Wiki + WMT & 42.6 & 51.6 & 36.7 & 55.6 & 30.2 & 20.3 & 63.1 & 40.9\\
 & 1-to-N Distill & 44.4 & 56.1 & 38.3 & 57.1 & 34.6 & 22.2 & 59.9 & 42.9\\
\midrule
rembert-12 & Fine-tuning & 43.8 & 57.4 & 36.7 & 56.1 & 33.0 & 23.4 & 62.2 & 37.5\\
 & N Fine-tuning & 40.6 & 58.1 & 34.1 & 50.6 & 35.0 & 22.7 & 48.4 & 35.4\\
 & Distill. WMT & 44.8 & 59.3 & 39.3 & 56.0 & 34.7 & 22.9 & 63.6 & 38.1\\
 & Distill. Wiki + WMT & 47.3 & 59.2 & 40.8 & 57.9 & 37.4 & 26.4 & 65.3 & 44.2\\
 & 1-to-N Distill & 48.4 & 64.2 & 40.2 & 57.6 & 41.3 & 28.4 & 63.7 & 43.5\\
\bottomrule
\end{tabular}
\caption{Segment-level agreement with human ratings on from-English language pairs. The metric is WMT Metrics DaRR \cite{mathur2020results}, a robust variant of Kendall Tau, higher is better. We average all the language pairs to English. The dagger$^\dagger{}$ indicates that the results were obtained from the WMT report. }
\label{table:distillation-results-ext2}
\end{table*}

We report system-level and segment-level performance of the compact metrics on the MWT Metrics shared task 2020, extending the performance analysis of the distilled models.

We re-implemented the WMT Metrics benchmark using data provided by the organizers. The results are consistent with the published version~\cite{mathur2020results} except for segment-level to-English pairs, marked with a dagger$^\dagger{}$ in the tables. We ran \BLEURT{}, \BLEURT-Tiny{}, \BLEURT{}-English WMT'20, \BLEURText{} ourselves. The first two are available online,\footnote{\url{https://github.com/google-research/bleurt}} the latter two were submitted to the WMT Metrics shared task 2020 and were obtained from the authors. We also report results for three state-of-the-art metrics: \COMET~\cite{rei2020comet}, \textsc{PRISM}~\cite{thompson-post-2020-automatic}, and \textsc{YiSi-1}~\cite{thompson-post-2020-automatic}, using the WMT Metrics report. We only report results for from-English pairs because the benchmark implementations are consistent for these. We also add the baseline \texttt{N Fine-tuning}, which describes the performance of fine-tuning the N models presented in Section~\ref{sec:languages-appendix} directly on WMT data.

As observed in the past~\cite{mathur2020tangled, mathur2020results} system- and segment-level correlations present very different outcomes: the teacher RemBERT-32 is outperformed by several other metrics on both \texttt{en-*} and  \texttt{*-en}, and the impact of the distillation improvements is mixed on to-English. A possible explanation is that system-level involves small sample sizes and that the data is very noisy~\cite{freitag2021experts}. Another explanation is that systems-level quality assessment is simply another task, which requires its own set of optimizations. In spite of these divergences, Table~\ref{table:distillation-results-ext4} shows that our contributions bring solid improvements on \texttt{en-*} (up to 20.2\%), which validates our approach.

\begin{table*}[!t]
\centering
\scriptsize
\begin{tabular} { l l | c | c c c c c c c c c c }
\toprule
 &   &  *-en &  cs-en &  de-en &  iu-en &  ja-en &  km-en &  pl-en &  ps-en &  ru-en &  ta-en &  zh-en\\
\midrule
\multicolumn{2}{l |}{BLEURT-Tiny~\cite{sellam2020bleurt}}    &  76.1 &  81.8 &  65.8 &  49.7 &  86.0 &  96.2 &  32.9 &  95.5 &  89.5 &  79.8 &  84.0\\
\multicolumn{2}{l |}{BLEURT~\cite{sellam2020bleurt}} & 76.2 &  73.2 &  81.3 &  53.4 &  81.4 &  96.6 &  30.7 &  94.4 &  82.6 &  76.4 &  92.1\\
\multicolumn{2}{l |}{BLEURT-English WMT'20~\cite{sellam2020learning}} &  74.9 &  72.5 &  77.0 &  32.0 &  82.0 &  98.4 &  37.1 &  95.5 &  84.4 &  76.8 &  93.1\\
\multicolumn{2}{l |}{BLEURT-Ext.~\cite{sellam2020learning}} & 73.1 &  66.8 &  81.8 &  35.9 &  77.2 &  98.5 &  29.8 &  94.2 &  79.7 &  74.3 &  93.1\\
\midrule
RemBERT-32 &    &  75.7 &  67.1 &  79.0 &  51.2 &  79.6 &  99.6 &  33.4 &  95.9 &  80.7 &  77.4 &  93.5\\
\midrule
Rembert-3 &  Fine-tuning &  77.4 &  78.4 &  74.2 &  53.9 &  90.3 &  96.9 &  27.0 &  92.3 &  86.1 &  80.8 &  94.0\\
 &  N Fine-tuning &  77.9 &  79.8 &  79.4 &  52.6 &  93.5 &  94.5 &  27.0 &  92.6 &  87.3 &  78.6 &  94.2\\
 &  Distill. WMT &  77.5 &  76.5 &  78.8 &  54.1 &  91.4 &  96.2 &  24.4 &  92.0 &  85.6 &  81.5 &  94.3\\
 &  Distill. Wiki + WMT &  77.3 &  80.8 &  76.3 &  54.8 &  91.5 &  97.6 &  23.0 &  89.0 &  84.5 &  80.3 &  95.3\\
 &  1-to-N Distill &  78.6 &  78.9 &  77.8 &  59.1 &  92.3 &  98.4 &  25.1 &  92.2 &  85.9 &  80.9 &  95.1\\
\midrule
Rembert-6 &  Fine-tuning &  77.8 &  80.0 &  76.9 &  65.4 &  88.3 &  95.2 &  20.5 &  91.1 &  85.0 &  81.2 &  94.2\\
 &  N Fine-tuning &  75.9 &  77.0 &  76.5 &  56.6 &  89.1 &  95.7 &  20.8 &  91.3 &  80.5 &  78.2 &  93.2\\
 &  Distill. WMT &  77.0 &  75.5 &  78.1 &  60.0 &  86.4 &  96.7 &  24.3 &  93.0 &  82.1 &  80.5 &  93.5\\
 &  Distill. Wiki + WMT &  78.2 &  80.1 &  76.7 &  60.8 &  88.9 &  98.9 &  21.6 &  93.6 &  85.5 &  81.1 &  94.9\\
 &  1-to-N Distill &  77.0 &  76.9 &  76.7 &  58.0 &  86.9 &  99.1 &  22.2 &  94.3 &  82.6 &  78.7 &  94.5\\
\midrule
Rembert-12 &  Fine-tuning &  76.7 &  77.4 &  77.7 &  59.8 &  85.5 &  95.3 &  22.9 &  91.7 &  82.9 &  80.2 &  93.4\\
 &  N Fine-tuning &  74.4 &  72.5 &  76.2 &  50.4 &  81.3 &  96.6 &  27.4 &  91.5 &  80.2 &  74.7 &  92.6\\
 &  Distill. WMT &  76.9 &  75.2 &  77.2 &  63.2 &  84.0 &  97.5 &  23.2 &  93.1 &  82.7 &  79.4 &  93.4\\
 &  Distill. Wiki + WMT &  77.4 &  76.3 &  75.7 &  62.1 &  87.3 &  99.3 &  22.1 &  94.2 &  84.0 &  78.4 &  94.1\\
 &  1-to-N Distill &  76.7 &  74.6 &  77.0 &  54.9 &  82.9 &  99.6 &  25.8 &  95.4 &  83.0 &  79.3 &  94.0\\
\bottomrule
\end{tabular}
\caption{System-level agreement with human ratings on to-English language pairs excluding outliers where they are available. The metric is Pearson correlation \cite{mathur2020results}, higher is better.}
\label{table:distillation-results-ext3}
\end{table*}

\begin{table*}[!t]
\centering
\scriptsize
\begin{tabular} { l l | c | c c c c c c c }
\toprule
 &   &  en-* &  en-cs &  en-de &  en-ja &  en-pl &  en-ru &  en-ta &  en-zh\\
\midrule
\multicolumn{2}{l |}{BLEURT-Ext.\cite{sellam2020learning}} &  90.3 &  96.0 &  87.0 &  95.3 &  82.8 &  98.0 &  81.4 &  91.5\\
\multicolumn{2}{l |}{COMET~\cite{rei2020comet}$^\dagger{}$} & 75.5 & 92.6 & 86.3 & 96.9 & 80.0 & 92.5 & 79.8 & 0.7\\
\multicolumn{2}{l |}{PRISM~\cite{thompson-post-2020-automatic}$^\dagger{}$} & 67.4 & 80.5 & 85.1 & 92.1 & 74.2 & 72.4 & 45.2 & 22.1\\
\multicolumn{2}{l |}{YiSi-1~\cite{thompson-post-2020-automatic}$^\dagger{}$} & 86.1 & 66.4 & 88.7 & 96.7 & 71.4 & 92.6 & 90.9 & 95.9\\
\midrule
RemBERT-32 &    &  83.6 &  96.1 &  86.2 &  97.1 &  85.4 &  91.4 &  80.8 &  48.6\\
\midrule
Rembert-3 &  Fine-tuning &  65.8 &  56.7 &  82.3 &  95.2 &  73.2 &  53.1 &  90.6 &  9.7\\
 &  N Fine-tuning &  73.6 &  67.8 &  80.9 &  92.7 &  67.1 &  76.5 &  88.6 &  41.7\\
 &  Distill. WMT &  69.2 &  69.0 &  79.0 &  94.1 &  75.6 &  62.4 &  88.3 &  16.2\\
 &  Distill. Wiki + WMT &  76.2 &  63.9 &  85.8 &  94.4 &  73.4 &  91.3 &  90.7 &  34.1\\
 &  1-to-N Distill &  78.9 &  67.9 &  86.6 &  93.8 &  73.7 &  94.7 &  89.5 &  46.3\\
\midrule
Rembert-6 &  Fine-tuning &  68.2 &  64.3 &  82.9 &  94.3 &  74.4 &  74.1 &  86.6 &  0.8\\
 &  N Fine-tuning &  81.8 &  75.3 &  84.3 &  91.4 &  79.4 &  91.9 &  87.8 &  62.8\\
 &  Distill. WMT &  70.3 &  68.8 &  84.4 &  93.7 &  75.5 &  69.7 &  87.7 &  12.7\\
 &  Distill. Wiki + WMT &  81.5 &  70.6 &  87.6 &  95.8 &  75.1 &  94.8 &  89.5 &  57.1\\
 &  1-to-N Distill &  82.0 &  75.8 &  86.1 &  93.1 &  76.2 &  96.0 &  83.6 &  63.4\\
\midrule
Rembert-12 &  Fine-tuning &  74.2 &  75.2 &  83.3 &  92.4 &  80.2 &  89.2 &  85.6 &  13.2\\
 &  N Fine-tuning &  77.1 &  85.4 &  86.2 &  88.6 &  85.7 &  85.4 &  85.1 &  23.1\\
 &  Distill. WMT &  72.3 &  75.9 &  84.7 &  93.3 &  79.3 &  72.4 &  85.3 &  14.9\\
 &  Distill. Wiki + WMT &  85.8 &  79.8 &  88.1 &  96.2 &  81.4 &  97.1 &  84.8 &  72.9\\
 &  1-to-N Distill &  86.3 &  87.9 &  87.8 &  94.0 &  82.0 &  96.9 &  82.2 &  73.5\\
\bottomrule
\end{tabular}
\caption{System-level agreement with human ratings on from-English language pairs, excluding outliers where they are available. The metric is Pearson correlation \cite{mathur2020results}, higher is better. The dagger$^\dagger{}$ indicates that the results were obtained from the WMT report. }
\label{table:distillation-results-ext4}
\end{table*}

\end{document}